\documentclass[10pt,twocolumn,letterpaper]{article}

\usepackage{cvpr}
\usepackage{times}
\usepackage{epsfig}
\usepackage{graphicx}
\usepackage{amsmath}
\usepackage{amssymb}
\usepackage{xcolor,soul,colortbl}
\usepackage{microtype}
\usepackage{flushend}
\usepackage{subcaption}
\usepackage{pdfpages}

\cvprfinalcopy %

\def\cvprPaperID{5150} %

\ifcvprfinal\fi
\begin{document}

\definecolor{revcolor}{RGB}{255,50,0}
\definecolor{rev2color}{RGB}{0,50,255}
\newcommand\rev[1] {\emph{\textcolor{revcolor}{#1}}}
\newcommand\REV[1] {\emph{\textcolor{rev2color}{#1}}}

\title{StyleRig: Rigging StyleGAN for 3D Control over Portrait Images}
\author{Ayush Tewari$^1$~~~Mohamed Elgharib$^1$~~~Gaurav Bharaj$^2$~~~Florian Bernard$^1$ \\ \vspace{0.4cm}
Hans-Peter Seidel$^1$~~~Patrick P{\'e}rez$^3$~~~Michael Zollh{\"o}fer$^4$~~~Christian Theobalt$^1$\\ 
		$^1$MPI Informatics, Saarland Informatics Campus~~~$^2$Technicolor~~~$^3$Valeo.ai~~~$^4$Stanford University
}
\twocolumn[{%
\renewcommand\twocolumn[1][]{#1}%
\maketitle
\begin{center}
\vspace{-0.5cm}
\includegraphics[width=\linewidth]{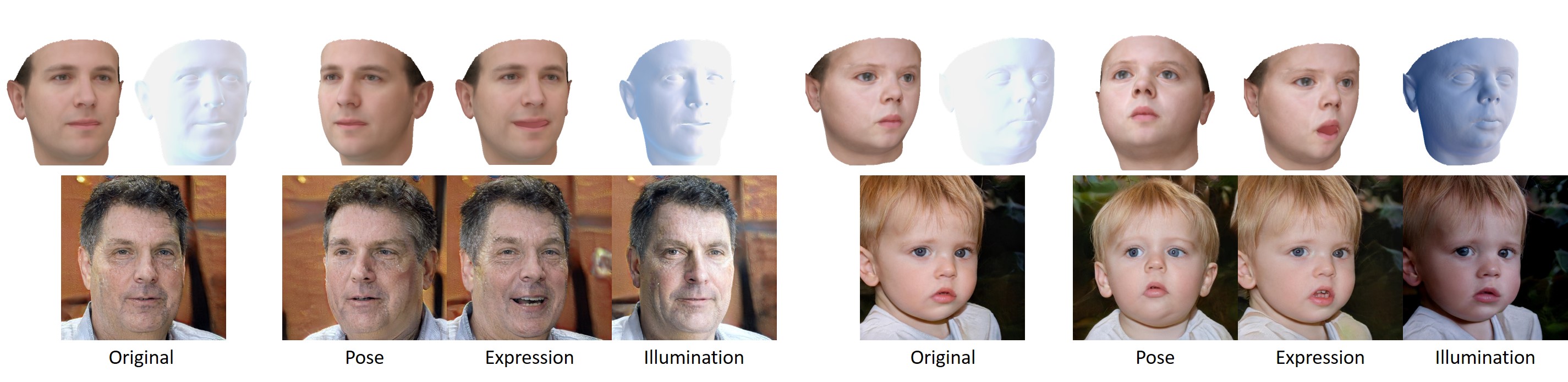}
\vspace{-0.8cm}
\captionof{figure}{
StyleRig allows for face rig-like control over StyleGAN generated portrait images, by
    translating semantic edits on 3D face meshes to the input space of StyleGAN.
 }
    \label{fig:editing}
	\end{center}
}]

\begin{abstract}
StyleGAN \cite{Karra19} generates photorealistic portrait images of faces with eyes, teeth, hair and context (neck, shoulders, background), but lacks a rig-like control over semantic face parameters that are interpretable in 3D, such as face pose, expressions, and scene illumination. Three-dimensional morphable face models (3DMMs) \cite{egger20193d} on the other hand offer control over the semantic parameters, but lack photorealism when rendered and only model the face interior, not other parts of a portrait image (hair, mouth interior, background). We present the first method to provide a face rig-like control over a pretrained and fixed StyleGAN via a 3DMM. A new rigging network, \textit{RigNet} is trained between the 3DMM's semantic parameters and StyleGAN's input. The network is trained in a self-supervised manner, without the need for manual annotations. At test time, our method generates portrait images with the photorealism of StyleGAN and provides explicit control over the 3D semantic parameters of the face.
\end{abstract}
\vspace{-0.2cm}

\section{Introduction}

Photorealistic synthesis of portrait face images finds many applications in several fields including special effects, extended reality, virtual worlds, and next-generation communication. During the content creation process for such applications, artist control over the \textit{face rig's} semantic parameters, such as geometric identity, expressions, reflectance, or scene illumination is desired. 
The computer vision and graphics communities have a rich history of modeling face rigs \cite{Li17,Richardson_2017_CVPR,RingNet19,Tewari19FML}. These models provide artist-friendly control (often called a face rig), while navigating the various parameters of a morphable face model (3DMM) \cite{Blanz2003,Blanz1999}. Such methods are often limited by the lack of training data, and more importantly, lack of photorealism in the final rendering.

Through 3D face scanning techniques high-quality face geometry datasets can be obtained \cite{BoothRPDZ18,Li17}. However, models derived from these datasets are bound by the diversity of faces scanned and may limit the generalization over the rich set of human faces' semantic parameterization. Further, deep learning-based models trained on \emph{in-the-wild} data \cite{Tewari19FML,tewari2018self,Tran19} also often rely on data-driven priors and other forms of regularization obtained from scan-based datasets. With respect to photorealism, perceptual losses recently showed an improvement of face modeling quality \cite{deng2019,Tran19} over existing methods. However, they still do not engender photorealistic face renders. Mouth interiors, hair, or eyes, let alone image background are often not modeled by such approaches.
Generative Adversarial Networks (GANs) \cite{NIPSGood} have lately achieved photorealism \cite{Isola17,KarraALL2018}, especially for faces. Karras~\etal~\cite{KarraALL2018} show that through a progressive growth of GAN's generator and discriminator, one can better stabilize and speed up training. When trained on the CelebA-HQ~\cite{KarraALL2018} dataset this yields a remarkable level of photorealism for faces. Their approach also shows how photorealistic face images of non-existent people can be sampled from the learned GAN distribution. 
Building on Karras~\etal\cite{KarraALL2018}, StyleGAN \cite{Karra19} uses ideas from the style transfer literature \cite{Gatys16,selim16} and proposes an architecture capable of disentangling various face attributes. Promising results of control over various attributes, including coarse (hair, geometry), medium (expressions, facial hair) and fine (color distribution, freckles) attributes were shown. 
However, these controllable attributes are not semantically well defined, and contain several similar yet entangled semantic attributes. For example, both coarse and medium level attributes contain face identity information. In addition, the coarse levels contain several entangled attributes such as face identity and  head pose.

We present a novel solution to \textit{rig} StyleGAN using a semantic parameter space for faces. Our approach brings the best of both worlds: the controllable parametric nature of existing morphable face models \cite{RingNet19,Tewari19FML}, and the high photorealism of generative face models \cite{KarraALL2018,Karra19}.
We employ a fixed and pretrained StyleGAN and do not require more data for training.
Our focus is to provide computer graphics style rig-like control over the various semantic parameters.
Our novel training procedure is based on a self-supervised two-way cycle consistency loss that is empowered by the combination of a face reconstruction network with a differentiable renderer.
This allows us to measure the photometric rerendering error in the image domain and leads to high quality results.
We show compelling results of our method, including interactive control of StyleGAN generated imagery as well as image synthesis conditioned on well-defined semantic parameters.%

\section{Related Work}
In the following, we discuss deep generative models for the synthesis of imagery with a focus on faces, as well as 3D parametric face models. %
For an in-depth overview of parametric face models and their possible applications we refer to the recent survey papers~\cite{egger20193d,Zollhoefer2018FaceSTAR}.

\paragraph{Deep Generative Models}
Generative adversarial networks (GANs) contain two main blocks: a generator and a discriminator \cite{NIPSGood}.
The generator takes a noise vector as an input and produces an output, and tries to fool the discriminator, whose purpose is to classify whether the output is real or fake.
When the input to the network is a noise vector, the output is a sample from the learned distribution.
Karras~\etal \cite{KarraALL2018} show that such a noise vector can generate high-resolution photorealistic images of human faces.
To achieve this they employ a progressive strategy of slowly increasing the size of the generator and the discriminator, by adding more layers during training.
This enables more stable training phase, and in turn helps learn high-resolution images of faces.
StyleGAN \cite{Karra19} can synthesize highly photorealistic images while allowing for more control over the output, compared to Karras~\etal\cite{KarraALL2018}. However, StyleGAN still suffers from a clear entanglement of semantically different attributes. Therefore, it does not provide a semantic and interpretable control over the image synthesis process. %
Exploring the latent space of GANs for image editing has been recently explored in Jahanian~\etal~\cite{gansteerability}. 
They can only achieve simple transformations, such as zoom and 2D translations as they need ground truth images for each transformation during training. 
For faces, concurrent efforts have been made in controlling images synthesized by GANs~\cite{Abdal_2019_ICCV,shen2019interpreting}, but they lack explicit rig-like 3D control of the generative model.
Isola \etal \cite{Isola17} use conditional GANs to produce image-to-image translations.
Here, the input is not a noise vector, but a conditional image from a source domain, which is translated to the target domain by the generator.
Their approach, however, requires paired training data.
CycleGAN \cite{CycleGAN2017} and UNIT \cite{Liu17UNIT} learn to perform image-to-image translation only using unpaired data using cycle-consistency losses. 
GAUGAN \cite{Park19SPADE} shows interactive semantic image synthesis based on spatially adaptive normalization.
The remarkable quality achieved by GANs has inspired the development of several neural rendering applications for faces \cite{egger20193d,Zollhoefer2018FaceSTAR,Tewari2020NeuralSTAR} and others objects \cite{chan2019dance,Brualla18,Yu_2019_CVPR}. 
\paragraph{3D Morphable Models}

3D Morphable Models (3DMMs) are commonly used to represent faces \cite{Blanz2003,Blanz1999}.
Here, faces are parameterized by the identity geometry, expressions, skin reflectance and scene illumination.
Expressions are commonly modeled using blendshapes, and illumination is generally modeled via spherical harmonics parameters~\cite{tewari17MoFA}.
The models are learned from 3D scans of people \cite{BoothRPDZ18,Li17}, or more recently from in-the-wild internet footage \cite{Tewari19FML}.
The parametric nature of 3DMMs allows navigating and exploring the space of plausible faces, e.g., in terms of geometry, expressions and so on.
Thus, synthetic images can be rendered based on different parameter configurations.
The rendered images, however, often look synthetic and lack photorealism.
More recently, neural rendering has been used to bridge the gap between synthetic computer graphics renderings and corresponding real versions \cite{kim2018DeepVideo,thies2019deferred,Usman_2019_ICCV,gecer2018semi}. 
Several methods have been proposed for fitting face models to images~\cite{Cao16,GarriZCVVPT2016,genova2018unsupervised,KimZTTRT17,Richardson_2017_CVPR,RingNet19,sela2017unrestricted,Tewari19FML,tewari2018self,tewari17MoFA,Tran19}.
Our work, however, focuses on learning-based approaches, that can be categorized into reconstruction only techniques~\cite{KimZTTRT17,Richardson_2017_CVPR,RingNet19,sela2017unrestricted,tewari17MoFA}, and reconstruction plus model learning~\cite{Tewari19FML,tewari2018self,Tran19}.
MoFA~\cite{tewari17MoFA} projects a face into the 3DMM space using a CNN, followed by a differentiable renderer to synthesize the reconstructed face.
The network is trained in a self-supervised manner based on a large collection of face images.
Tran~\etal \cite{Tran19} use a perceptual loss to enhance the renderings of the reconstruction.
RingNet~\cite{RingNet19} and FML~\cite{Tewari19FML} impose multi-image consistency losses to enforce identity similarity.
RingNet also enforces identity dissimilarity between pictures of different people.
Several approaches learn to reconstruct the parameters of a 3DMM by training it on large scale synthetic data~\cite{KimZTTRT17,Richardson_2017_CVPR,sela2017unrestricted}.
For a more comprehensive overview of all techniques please refer to \cite{egger20193d,Zollhoefer2018FaceSTAR}.

\section{Overview}
StyleGAN \cite{Karra19} can be seen as a function that maps a latent code $\mathbf{w} \in \mathbb{R}^{l}$ to a realistic portrait image $\mathbf{I}_\mathbf{w} = \textit{StyleGAN}(\textbf{w}) \in \mathbb{R}^{3\times w\times h}$ of a human face.
While the generated images are of very high quality and at a high resolution ($w = h = 1024$), there is no semantic control over the generated output, such as the head pose, expression, or illumination.
StyleRig allows us to obtain a rig-like control over StyleGAN-generated facial imagery in terms of semantic and interpretable control parameters (Sec.~\ref{sec:results}).
In the following, we explain the semantic control space (Sec.~\ref{sec:control}), training data (Sec.~\ref{sec:corpus}), network architecture (Sec.~\ref{sec:network}) and loss function (Sec.~\ref{sec:loss}).

\section{Semantic Rig Parameters} \label{sec:control}
Our approach uses a parametric face model to achieve an explicit rig-like control of StyleGAN-generated imagery based on a set of semantic control parameters.
The control parameters are a subset of $\mathbf{p} = (\boldsymbol\alpha, \boldsymbol\beta, \boldsymbol\delta, \boldsymbol\gamma, \mathbf{R}, \mathbf{t}) \in\mathbb{R}^{f}$, which describes the facial shape $\boldsymbol\alpha\in\mathbb{R}^{80}$, skin reflectance $\boldsymbol\beta\in\mathbb{R}^{80}$, facial expression $\boldsymbol\delta\in\mathbb{R}^{64}$, scene illumination $\boldsymbol\gamma\in\mathbb{R}^{27}$, head rotation $\mathbf{R}\in SO(3)$, and translation $\mathbf{t}\in\mathbb{R}^3$, with the dimensionality of $\mathbf{p}$ being $f = 257$. %
We define the control space for the facial shape $\boldsymbol\alpha$ and skin reflectance $\boldsymbol\beta$ using two low-dimensional affine models that have been computed via Principal Component Analysis (PCA) based on $200$ ($100$ male, $100$ female) scans of human faces \cite{Blanz1999}.
The output of this model is represented by a triangle mesh with $53k$ vertices and per-vertex color information.
The control space for the expression $\boldsymbol\delta$ is given in terms of an additional affine model that captures the expression dependent displacement of the vertices.
We obtain this model by applying PCA to a set of blendshapes \cite{Alexander:2009,Cao:2014} which have been transferred to the topology of the shape and reflectance models.
The affine models for shape, appearance, and expression cover more than 99\% of the variance in the original datasets.
Illumination $\boldsymbol\gamma$ is modeled based on three bands of spherical harmonics per color channel leading to an additional $27$ parameters.

\section{Training Corpus} \label{sec:corpus}
Besides the parametric face model, our approach requires a set of face images $\mathbf{I}_{\mathbf{w}}$ and their corresponding latent codes $\mathbf{w}$ as training data.
To this end, we sample $N=200$k latent codes $\mathbf{w} \in \mathbb{R}^l$ and generate the corresponding photorealistic face images $\mathbf{I}_{\mathbf{w}} = \textit{StyleGAN}(\mathbf{w})$ using a pretrained StyleGAN network.
We use $l=18\times512$ dimensional latent space which is the output of the mapping network in StyleGAN, as it has been shown to be more disentangled~\cite{Abdal_2019_ICCV,Karra19}.
Here, $18$ latent vectors of size $512$ are used at different resolutions. 
Each training sample is generated by combining up to $5$ separately sampled latent vectors, similar to the mixing regularizer in Karras~\etal~\cite{Karra19}.
This allows our networks to reason independently about the latent vectors at different resolutions. 
Given these $(\mathbf{w}, \mathbf{I}_{\mathbf{w}})$ pairs, our approach can be trained in a self-supervised manner without requiring any additional image data or manual annotations.

\begin{figure*}
    \centering
    \includegraphics[width=0.95\linewidth]{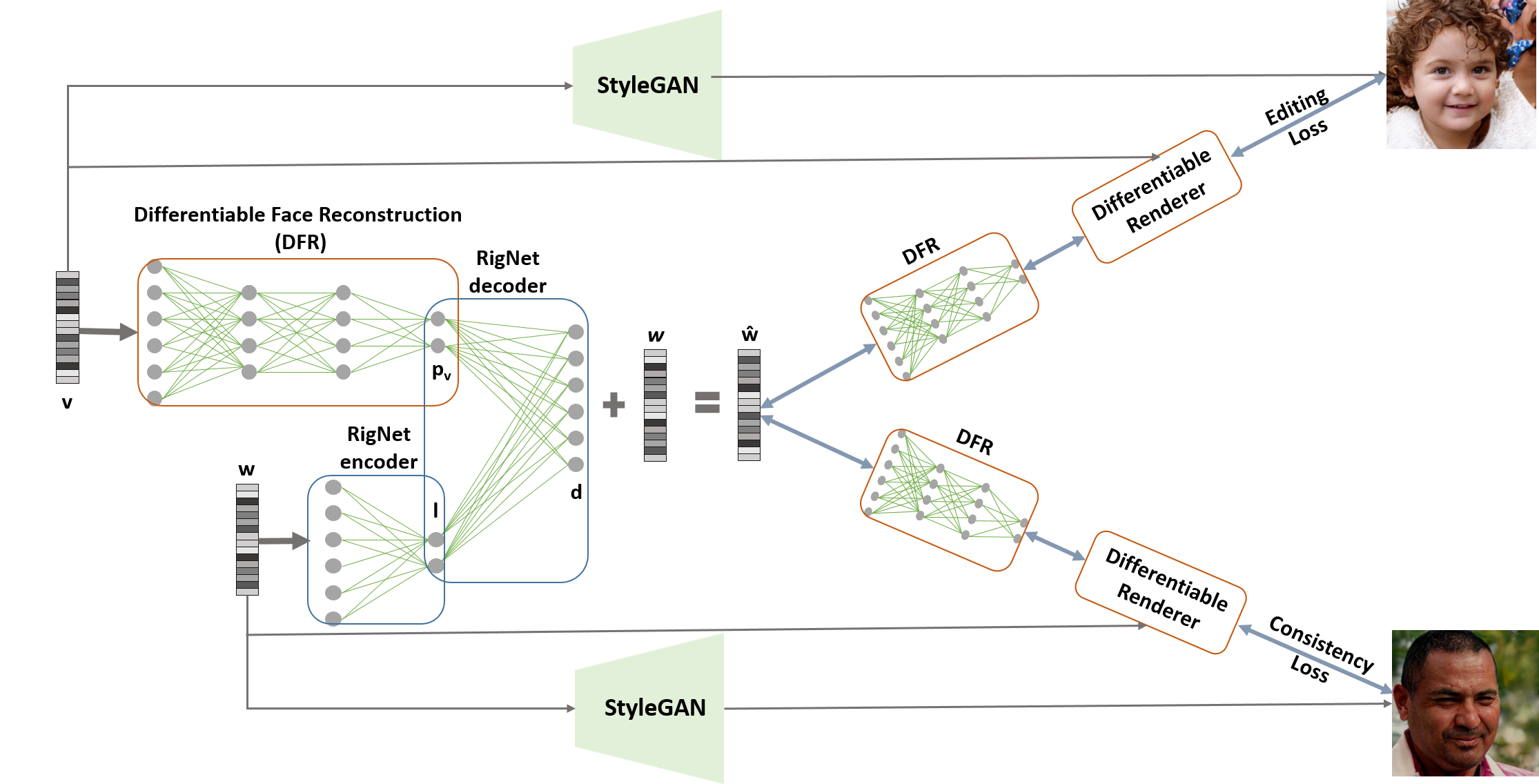}
    \caption{
    StyleRig enables rig-like control over StyleGAN-generated facial imagery based on a learned rigger network (RigNet).
    To this end, we employ a self-supervised training approach based on a differentiable face reconstruction (DFR) and a neural face renderer (StyleGAN).
    The DFR and StyleGAN networks are pretrained and their weights are fixed, only RigNet is trainable.
    We define the consistency and edit losses in the image domain using a differentiable renderer.
    }
    \label{fig:pipeline}
\end{figure*}

\begin{figure}
    \centering
    \includegraphics[width=\linewidth]{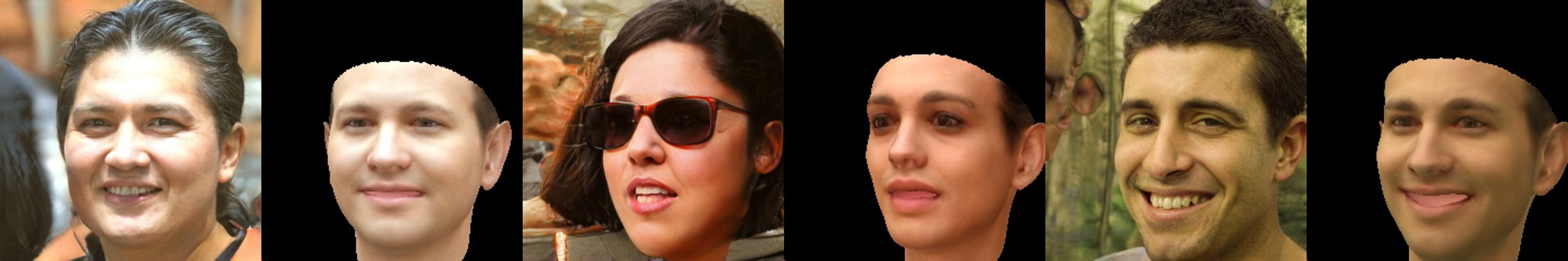}
    \vspace{-0.5cm}
    \caption{Differentiable Face Reconstruction. Visualized are (image, reconstruction) pairs. The network however, only gets the latent vector corresponding to the images as input.}
    \label{fig:DFR}
\end{figure}

\section{Network Architecture} \label{sec:network}
Given a latent code $\mathbf{w}\in\mathbb{R}^l$ that corresponds to an image $\mathbf{I}_\mathbf{w}$, and a vector $\mathbf{p}\in\mathbb{R}^f$ of semantic control parameters, we want to learn a function that outputs a modified latent code $\mathbf{\hat w}=\textit{RigNet}(\mathbf{w},\mathbf{p})$.
The modified latent code $\mathbf{\hat w}$ should map to a modified face image $\mathbf{I}_{\mathbf{\hat w}}=\textit{StyleGAN}(\mathbf{\hat w})$ that obeys the control parameters $\mathbf{p}$.
One example would be changing the rotation of the face in an image such that it matches a given target rotation, while maintaining the facial identity, expression, and scene illumination (see Sec.~\ref{sec:results} for examples).
We train separate RigNet networks for the different modes of control i.e., pose, expressions and illumination. 
RigNet is implemented based on a linear two-layer perceptron (MLP).
We propose a self-supervised training of RigNet based on two-way cycle consistency losses and a differentiable face reconstruction (DFR) network. Fig.~\ref{fig:pipeline} shows an overview of our architecture.
Our network combines several components that fulfill specific tasks.

\textbf{Differentiable Face Reconstruction}
One key component is a pretrained differentiable face reconstruction (DFR) network.
This parameter regressor is a function $\mathcal{F}:\mathbb{R}^{l} \rightarrow \mathbb{R}^{f}$ that maps a latent code $\mathbf{w}$ to a vector of semantic control parameters $\mathbf{p}_{\mathbf{w}}=\mathcal{F}(\mathbf{w})$.
In practice, we model $\mathcal{F}$ using a three layer MLP with ELU activations after every intermediate layer, and train it in a self-supervised manner.
This requires a differentiable render layer $\mathcal{R}: \mathbb{R}^{f} \rightarrow \mathbb{R}^{3\times w \times h}$ that takes a face parameter vector $\mathbf{p}$ as input, converts it into a 3D mesh and generates a synthetic rendering $\mathbf{S}_{\mathbf{w}} = \mathcal{R}(\mathbf{p}_{\mathbf{w}})$ of the face\footnote{We use point-based rendering of the mesh vertices.}.
We then train $\mathcal{F}$ using a rerendering loss:
\begin{equation}
\label{eq:renderloss}
\mathcal{L}_{\text{render}}(\mathbf{I}_\mathbf{w}, \mathbf{p}) = \mathcal{L}_{\text{photo}}(\mathbf{I}_\mathbf{w}, \mathbf{p}) + \lambda_{\text{land}} \mathcal{L}_{\text{land}}(\mathbf{I}_\mathbf{w}, \mathbf{p})
\enspace{.}
\end{equation}
The first term is a dense photometric alignment loss:
$$
\mathcal{L}_{\text{photo}}(\mathbf{I}_\mathbf{w}, \mathbf{p}) = \big\| \mathbf{M} \odot (\mathbf{I}_\mathbf{w} - \mathcal{R}\big(\mathbf{p})\big)\big) \big\|_2^2 \enspace{.}
$$
Here, $\mathbf{M}$ is a binary mask with all pixels where the face mesh is rendered set to $1$ and $\odot$ is element-wise multiplication.
We also use a sparse landmark loss
$$
\mathcal{L}_{\text{land}}(\mathbf{I}_\mathbf{w}, \mathbf{p}) = \big\| \mathbf{L_{I_w}} - \mathbf{L_M} \big\|_2^2 \enspace{,}
$$
where $ \mathbf{L_{I_w}} \in \mathbb{R}^{66\times2}$ are $66$ automatically computed landmarks~\cite{SaragLC2011} on the image $\mathbf{I}_\mathbf{w}$, and $\mathbf{L_M}$ are the corresponding landmark positions on the rendered reconstructed face. 
The landmark vertices on the mesh are manually annoted before training.
$\lambda_\text{land}$ is a fixed weight used to balance the loss terms. 
In addition, we also employ statistical regularization on the parameters of the face model, as done in MoFA.~\cite{tewari17MoFA}.
After training, the weights of $\mathcal{F}$ are fixed.
Fig.~\ref{fig:DFR} shows some results of the reconstructions obtained by DFR. 

\textbf{RigNet Encoder}
The encoder takes the latent vector $\mathbf{w}$ as input and linearly transforms it into a lower dimensional vector $\mathbf{l}$ of size $18 \times 32$. 
Each sub-vector $\mathbf{w}_i$ of $\mathbf{w}$ of size $512$ is independently transformed into a sub-vector $\mathbf{l}_i$ of size $32$, for all $i\in \{0,\ldots,17\}$.

\textbf{RigNet Decoder}
The decoder tranforms $\mathbf{l}$ and the input control parameters $\mathbf{p}$ into the output $\hat{\mathbf{w}}$.
Similar to the encoder, we use independent linear decoders for each $\mathbf{l}_i$.
Each layer first concatenates $\mathbf{l}_i$ and $\mathbf{p}$, and transforms it into ${\mathbf{d}}_i$, for all $ i \in \{0,\ldots,17\}$.
The final output is computed as $\hat{\mathbf{w}} = \mathbf{d} + \mathbf{w}$.

\begin{figure*}
    \centering
    \includegraphics[width=\linewidth]{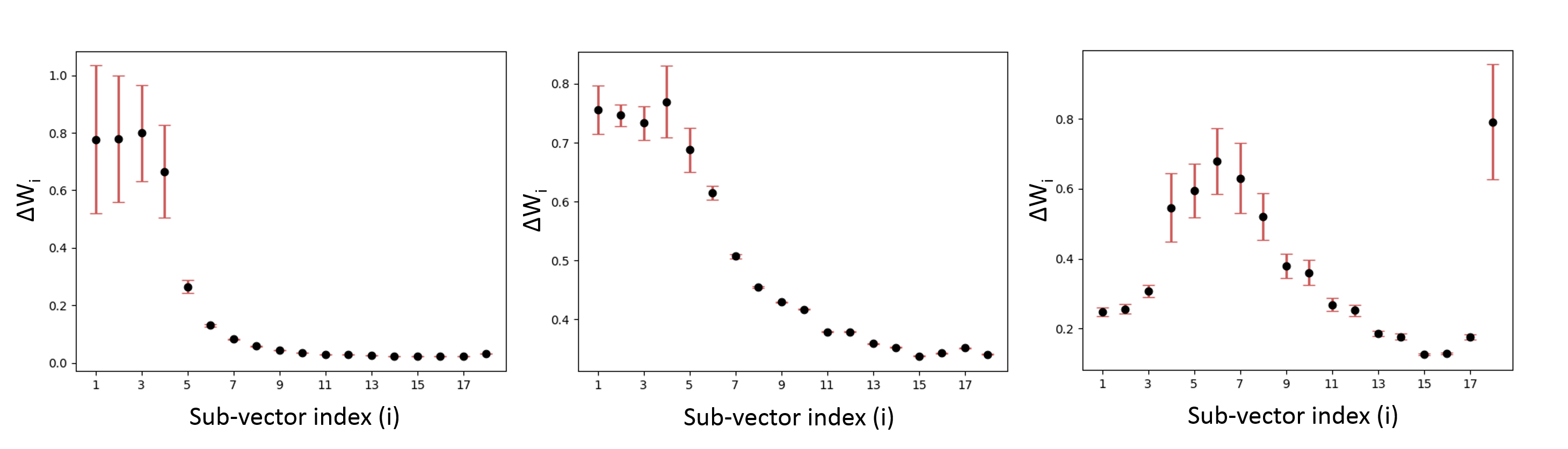}
    \vspace{-0.6cm}
    \caption{Change of latent vectors at different resolutions. Coarse vectors are responsible for rotation (left), medium for expressions (middle), medium and fine for illumination (right).
    }
    \vspace{-0.3cm}
    \label{fig:plot_mixing}
\end{figure*}

\section{Self-supervised Training} \label{sec:loss}
Our goal is to train RigNet such that we can inject a subset of parameters into a given latent code $\mathbf{w}$.
For example, we might want to inject a new head pose, while maintaining the facial identity, expression, and illumination in the original image synthesized from $\mathbf{w}$.
We employ the following loss function for training:
\begin{equation}
\label{eq:loss}
\mathcal{L}_{\text{total}} = \mathcal{L}_{\text{rec}} +  \mathcal{L}_{\text{edit}} +  \mathcal{L}_{\text{consist}} \enspace{.}
\end{equation}
It consists of a reconstruction loss $\mathcal{L}_{\text{rec}}$, an editing loss $\mathcal{L}_{\text{edit}}$, and a consistency loss $\mathcal{L}_{\text{consist}}$.
Since we do not have ground truth for the desired modifications (our training corpus only contains one image per person), we employ self-supervision based on cycle-consistent editing and consistency losses.
We optimize $\mathcal{L}_{\text{total}}$ based on AdaDelta~\cite{zeiler2012adadelta} with a learning rate of $0.01$.
In the following, we provide details.

\paragraph{Reconstruction Loss}
We want to design RigNet such that it reproduces the latent codes in the training corpus.
Formally, we want that $\mathit{RigNet}(\mathbf{w},\mathcal{F}(\mathbf{w})) = \mathbf{w}$.
We enforce this with the following $\ell_2$-loss:
$$
\mathcal{L}_{\text{rec}} = \big\|\mathit{RigNet}(\mathbf{w},\mathcal{F}(\mathbf{w})) - \mathbf{w}\big\|_2^2 \enspace{.}
$$
This constraint anchors the learned mapping at the right location in the latent space.
Without this constraint, learning the mapping is underconstrained, which leads to a degradation in the image quality (see Sec.~\ref{sec:results}).
Since $\mathcal{F}$ is pretrained and not updated, the semantics of the control space are enforced.

\paragraph{Cycle-Consistent Per-Pixel Editing Loss}
Given two latent codes, $\mathbf{w}$ and $\mathbf{v}$ with corresponding images $\mathbf{I}_{\mathbf{w}}$ and $\mathbf{I}_{\mathbf{v}}$, we transfer the semantic parameters of $\mathbf{v}$ to $\mathbf{w}$ during training.
We first extract the target parameter vector $\mathbf{p}_{\mathbf{v}}=\mathcal{F}(\mathbf{v})$ using the differentiable face reconstruction network. %
Next, we inject a subset of the parameters of $\mathbf{p}_\mathbf{v}$ (the ones we want to modify) into the latent code $\mathbf{w}$ to yield a new latent code $\mathbf{\hat w}=\mathit{RigNet}(\mathbf{w}, \mathbf{p}_\mathbf{v})$, so that $\mathbf{I}_{\mathbf{\hat w}} = \textit{StyleGAN}(\mathbf{\hat w})$ 
(ideally) corresponds to the image $\mathbf{I}_{\mathbf{w}}$, modified according to the subset of the parameters of $\mathbf{p}_\mathbf{v}$.
For example, $\mathbf{\hat w}$ might retain the facial identity, expression and scene illumination of $\mathbf{w}$, but should perform the head rotation specified in $\mathbf{p}_\mathbf{v}$.

Since we do not have ground truth for such a modification, i.e., the image $\mathbf{I}_{\mathbf{\hat w}}$ is unknown, we employ supervision based on a cycle-consistent editing loss.
The editing loss enforces that the latent code $\mathbf{\hat w}$ contains the modified parameters.
We enforce this by mapping from the latent to the parameter space $\mathbf{\hat p} = \mathcal{F}(\mathbf{\hat w})$.
The regressed parameters $\mathbf{\hat p}$ should have the same rotation as $\mathbf{p}_\mathbf{v}$.
We could measure this directly in the parameter space but this has been shown to not be very effective~\cite{tewari17MoFA}.
We also observed in our experiments that minimizing a loss in the parameter space does not lead to desired results, since the perceptual effect of different parameters in the image space can be very  different.

Instead, we employ a rerendering loss similar to the one used for differentiable face reconstruction.
We take the original target parameter vector $\mathbf{p}_{\mathbf{v}}$ and replace its rotation parameters with the regressed rotation from $\mathbf{\hat p}$, resulting in $\mathbf{p}_{\text{edit}}$.
We can now compare this to $\mathbf{I}_\mathbf{v}$ using the rerendering loss (see Eq.~\ref{eq:renderloss}):
$$
\mathcal{L}_\text{edit} = \mathcal{L}_{\text{render}}(\mathbf{I}_\mathbf{v}, \mathbf{p}_{\text{edit}})
\enspace{.}
$$
We do not use any regularization terms here. 
Such a loss function ensures that the rotation component of $\mathbf{p}_{\text{edit}}$ aligns with $\mathbf{I}_\mathbf{v}$, which is the desired output. 
The component of $\mathbf{p}_{\mathbf{v}}$ which is replaced from $\mathbf{\hat p}$ depends on the property we want to change. 
It could either be the pose, expressions, or illumination parameters.

\paragraph{Cycle-consistent Per-pixel Consistency Loss}
In addition to the editing loss, we enforce consistency of the parameters that should not be changed by the performed edit operation.
The regressed parameters $\mathbf{\hat p}$ should have the same unmodified parameters as $\mathbf{p}_\mathbf{w}$.
Similarly as above, we impose this in terms of a rerendering loss.
We take the original parameter vector $\mathbf{p}_{\mathbf{w}}$ and replace all parameters that should not be modified by the regressed ones from $\mathbf{\hat p}$, resulting in $\mathbf{p}_{\text{consist}}$.
In the case of modifying rotation values, the parameters that should not change are expression, illumination as well as identity parameters (shape and skin reflectance).
This leads to the loss function:
$$
\mathcal{L}_\text{consist} = \mathcal{L}_{\text{render}}(\mathbf{I}_\mathbf{w}, \mathbf{p}_{\text{consist}})
\enspace{.}
$$

\textbf{Siamese Training}
Since we have already sampled two latent codes $\mathbf{w}$ and $\mathbf{v}$ during training, we perform the same operations in a reverse order, i.e., in addition to injecting $\mathbf{p}_\mathbf{v}$ into $\mathbf{w}$, we also inject $\mathbf{p}_\mathbf{w}$ into $\mathbf{v}$.
To this end, we use a Siamese network with two towers that have shared weights.
This results in a two-way cycle consistency loss.

\begin{figure*}[!h]
    \centering
    \includegraphics[width=\linewidth]{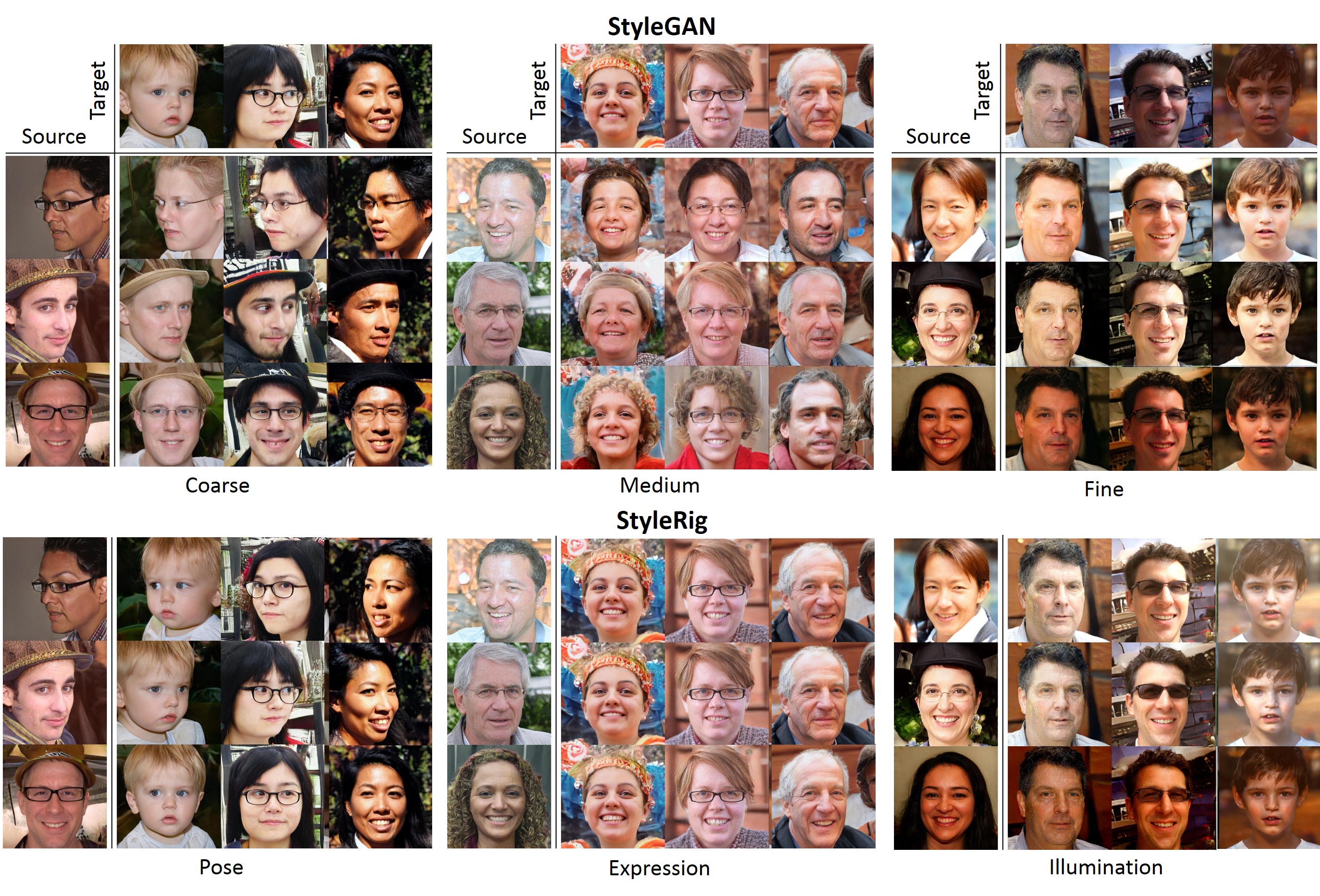}
    \vspace*{-0.6cm}
    \caption{Mixing between source and target images generated by StyleGAN. 
    For StyleGAN, the latent vectors of the source samples (rows) are copied to the target vectors (columns).
    StyleRig allows us to mix semantically meaningful parameters, i.e., head pose, expressions and scene illumination. These parameters can be copied over from the source to target images.}
    \label{fig:mixing}
    \vspace{-0.2cm}
\end{figure*}
\section{Results} \label{sec:results}
At test time, StyleRig allows control over the pose, expression, and illumination parameters of StyleGAN generated images.
We demonstrate the efficacy of our approach with three applications: Style Mixing (\ref{sec:stylemixing}), Interactive Rig Control (\ref{sec:interaction}) and Conditional Image Generation (\ref{sec:conditional}). 
\subsection{Style Mixing}
\label{sec:stylemixing}
Karras~\etal~\cite{Karra19} show StyleGAN vectors at different scales that correspond to different \emph{styles}.
To demonstrate \emph{style mixing}, latent vectors at certain resolutions are copied from a source to a target image, and new images are generated. 
As shown in Fig.~\ref{fig:mixing}, coarse styles contain information about the pose as well as identity, medium styles include information about expressions, hair structure, and illumination, while fine styles include the color scheme of the source.
We show a similar application of mixing, but with significantly more complete %
control over the semantic parameters.
To generate images with a target identity, we transfer the source parameters of our face rig to the target latent, resulting in images with different head poses, expressions and illumination.
This rig-like control is not possible via the mixing strategy of Karras~\etal which entangles multiple semantic dimensions in the mixed results. 
In Fig.~\ref{fig:plot_mixing}, we analyze how the latent vectors of StyleGAN are transformed by StyleRig. 
The figure shows the average change and variance (change is measured as $\ell_2$ distance) of StyleGAN latent vectors at all resolutions, computed over $2500$ mixing results. 
As expected, coarse latent code vectors are mainly responsible for  rotation.
Expression is controlled both by coarse and medium level latent codes.
The light direction is mostly controlled by the medium resolution vectors.
However, the fine latent vector also plays an important role in the control of the global color scheme of the images.
Rather than having to specify which vectors need to change and by how much, StyleRig recovers this mapping in a self-supervised manner.
As shown in Fig.~\ref{fig:mixing}, we can also preserve scene context like background, hair styles and accessories better.

\subsection{Interactive Rig Control}
\label{sec:interaction}
Since the parameters of the 3DMM can also be controlled independently, StyleRig allows for explicit semantic control of StyleGAN generated images.
We develop a user interface where a user can interact with a face mesh by interactively changing its pose, expression, and scene illumination parameters. %
These updated parameters are then %
fed into RigNet to generate new images at interactive frame rates ($\sim5$ fps).
Fig.~\ref{fig:editing} shows the results for various controls over StyleGAN images: pose, expression, and illumination edits. 
The control rig carries out the edits in a smooth interactive manner.
Please refer to the supplemental video for more results. 

\textbf{Analysis of StyleRig}
\begin{figure}
    \centering
    \includegraphics[width=\linewidth]{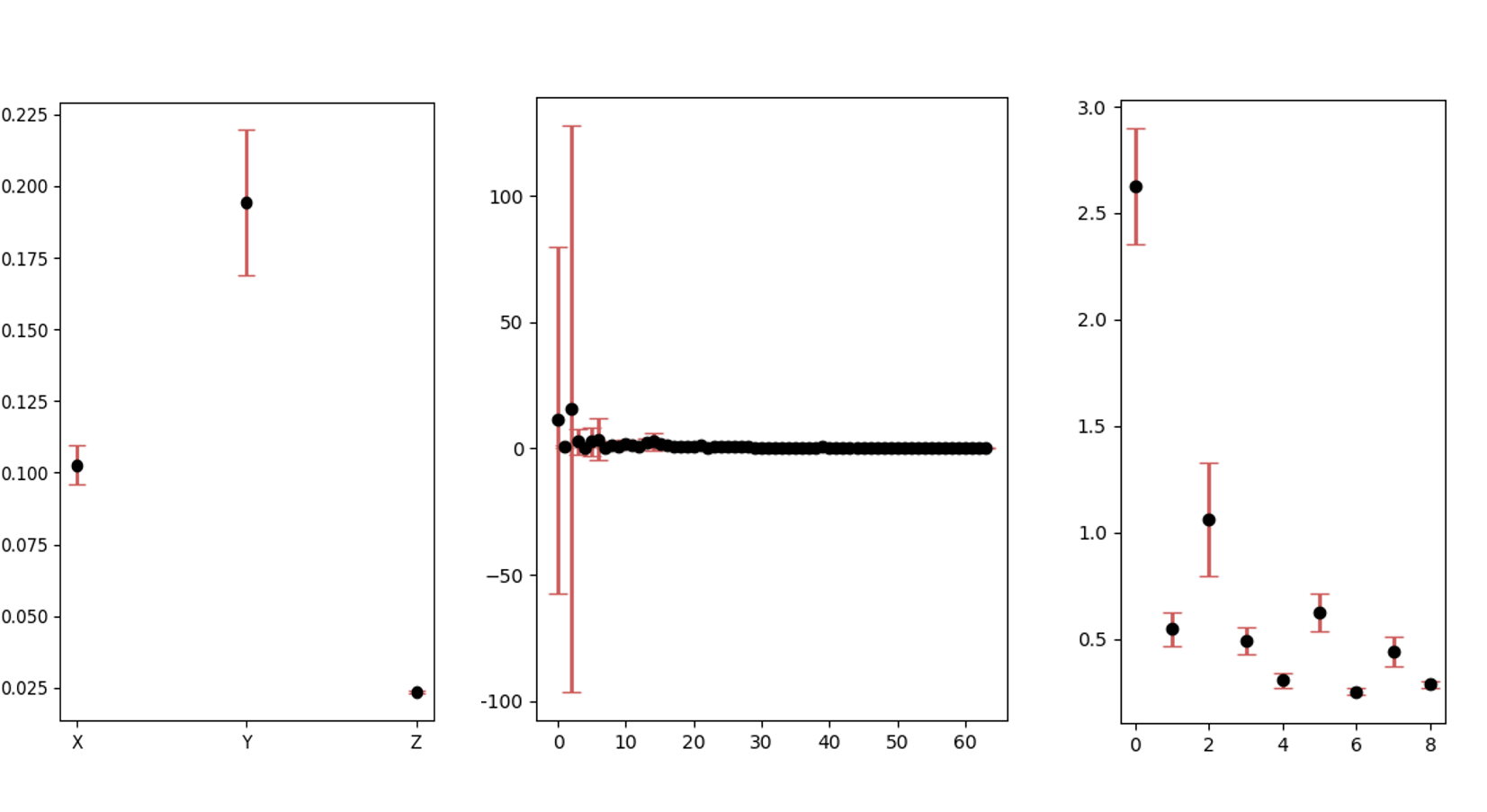}
    \vspace{-0.6cm}
    \caption{Distribution of face model parameters in the training data. \textit{x}-axis shows the face model parameters for rotation, expression and illumination from left-right. \textit{y}-axis shows the mean and variance of the parameters computed over $20k$ training samples.}
    \label{fig:training_data_dist}
\end{figure}
The interactive editor allows us to easily \emph{inspect} the trained networks. 
We observe that while the network does a good job at most controls, some expressivity of the 3D parametric face model is lost. That is, RigNet cannot transfer all modes of parametric control to similar changes in the StyleGAN generated images. 
For example, we notice that in-plane rotation of the face mesh is ignored.
Similarly, many expressions of the face mesh do not translate well into the resultant generated images.
We attribute these problems to the bias in the images StyleGAN has been trained on.
To analyze these modes, we look at the distribution of face model parameters in our training data, generated from StyleGAN, see Fig.~\ref{fig:training_data_dist}.
We notice that in-plane rotations (rotation around the \emph{Z}-axis) are hardly present in the data. In fact, most variation is only around the \emph{Y}-axis.
This could be because StyleGAN is trained on the Flickr-HQ dataset~\cite{Karra19}.
Most static images of faces in such a dataset would not include in-plane rotations.
The same reasoning can be applied to expressions, where most generated images consist of either neutral or smiling/laughing faces.
These expressions can be captured using up to three blendshapes.
Even though the face rig contains $64$ vectors, we cannot control them well because of the biases in the distribution of the training data.
Similarly, the lighting conditions are also limited in the dataset. We note that there are larger variations in the global color and azimuth dimensions, as compared to the other dimensions. 
Our approach provides an intuitive and interactive user interface which allows us to inspect not only StyleRig, but also the biases present in StyleGAN. 
\subsection{Conditional Image Generation}
\label{sec:conditional}
\begin{figure}
    \centering
    \includegraphics[width=\linewidth]{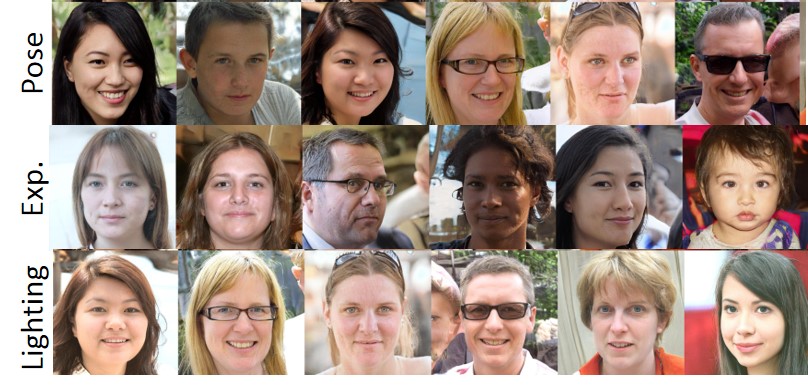}
    \vspace{-0.6cm}
    \caption{Explicit control over the 3D parameters allows us to turn StyleGAN into a conditional generative model. 
    }
    \label{fig:conditional}
\end{figure}

Explicit and implicit control of a pretrained generative model allows us to turn it into a conditional one. 
We can simply fix the pose, expression, or illumination inputs to RigNet in order to generate images which correspond to the specified parameters, see Fig.~\ref{fig:conditional}. 
This is a straight forward way to convert an unconditional generative model into a conditional model, and can produce high-resolution photorealistic results. 
It is also very efficient, as it takes us less than $24$ hours to train StyleRig, while training a conditional generative model from scratch should take at least as much time as StyleGAN, which takes more than $41$ days to train (both numbers are for an Nvidia Volta GPU).

\begin{figure}
    \centering
    \includegraphics[width=1\linewidth]{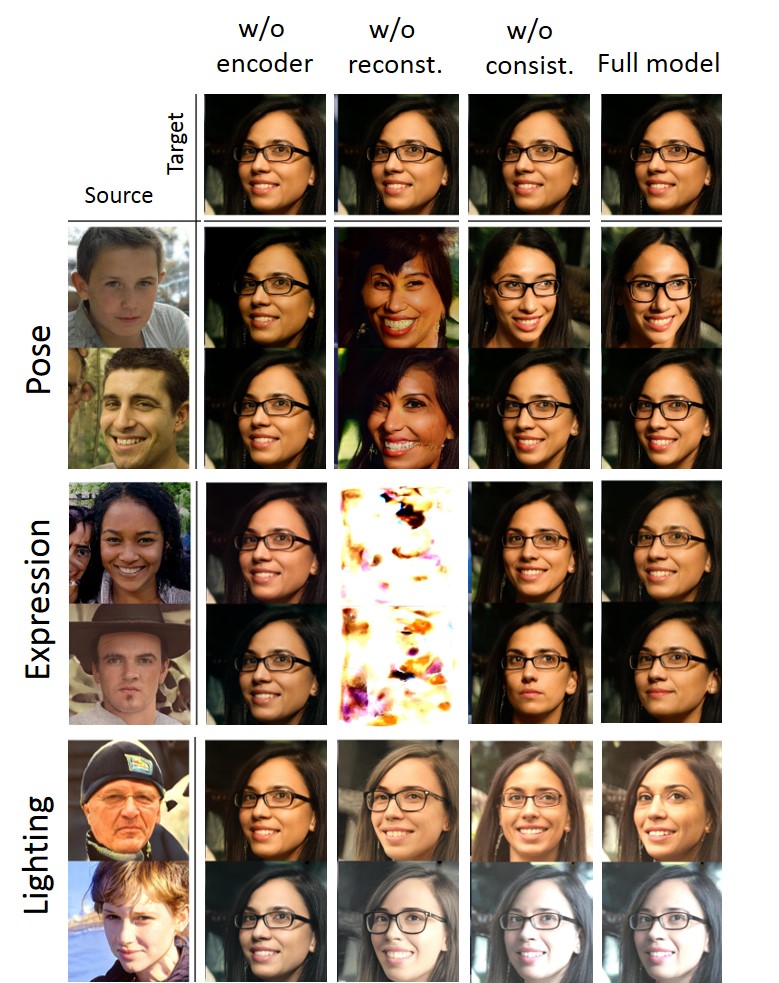}
    \vspace{-0.6cm}
    \caption{Baseline comparisons. Our full approach obtains the highest quality results.}
    \vspace{-0.2cm}
    \label{fig:baselines}
\end{figure}

\subsection{Comparisons to Baseline Approaches}
In the following, we compare our approach with several baseline approaches.

\textbf{``Steering'' the latent vector}
Inspired by Jahanian~\etal~\cite{gansteerability}, we design a network architecture which tries to \emph{steer} the StyleGAN latent vector based on the change in parameters.
This network architecture does not use the latent vector $\mathbf{w}$ as an input, and thus does not require an encoder. 
The inputs to the network are the delta in the face model parameters, with the output being the delta in the latent vector. 
In our settings, such an architecture does not lead to desirable results with the network not being able to deform the geometry of the faces, see Fig.~\ref{fig:baselines}.
Thus, the semantic deltas in latent space should also be conditional on the the latent vectors, in addition to the target parameters.

\textbf{Different Loss Functions}
As explained in Eq.~\ref{eq:loss}, our loss function consists of three terms.
For the first baseline, we switch off the reconstruction loss. 
This can lead to the output latent vectors drifting from the space of StyleGAN latent codes, thus resulting in non-face images.
Next, we switch off the consistency loss. 
This loss term enforces the consistency of all face model parameters, other than the one being changed. 
Without this term, changing one dimension, for example the illumination, also changes others such as the head pose. 
Our final model ensures the desired edits with consistent identity and scene information. Note that switching off the editing loss is not a good baseline, as it would not add any control over the generator.

\subsection{Simultaneous Parameter Control}
\begin{figure}
    \centering
    \includegraphics[width=\linewidth]{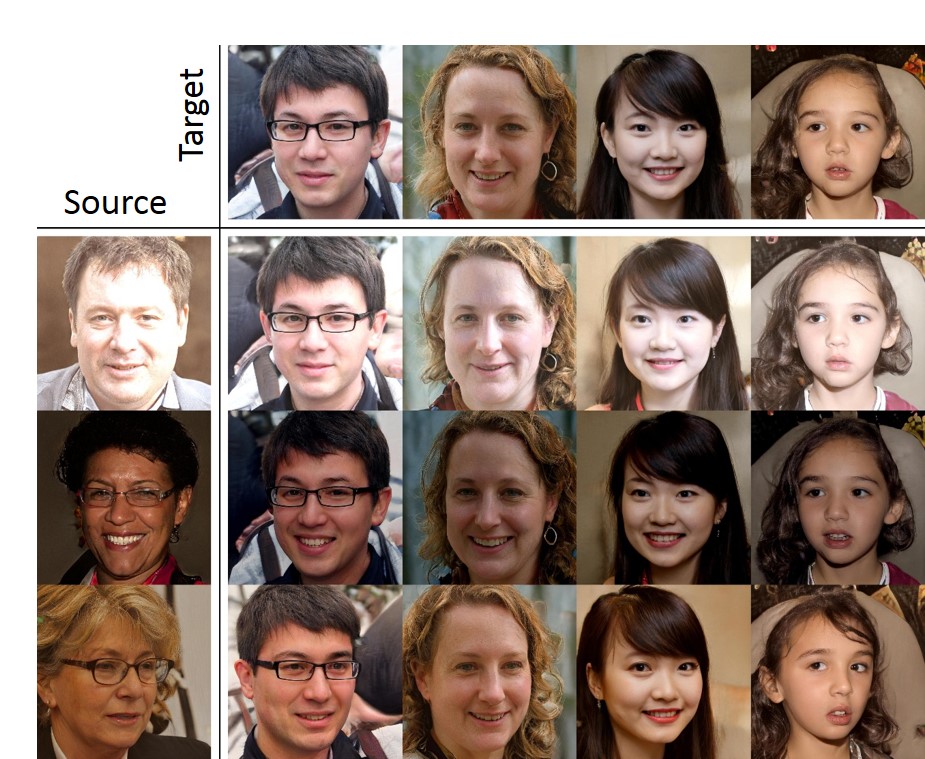}
    \vspace{-0.6cm}
    \caption{RigNet can also control pose, expression, and illumination parameters simultaneously. 
    These parameters are transferred from source to target images, while the identity in the target images is preserved.}
    \vspace{-0.3cm}
    \label{fig:mixing_all}
\end{figure}
In addition to controlling different parameters independently, we can also control them simultaneously. 
To this end, we train RigNet, such that, it receives target pose, expression, and illumination parameters as input. 
For every $(\mathbf{w}, \mathbf{v})$ training code vector pair, we sample three training samples. Here, one out of the three parameters (pose, expression or illumination) is changed in each sample.
We then use the loss function defined in Eq.~\ref{eq:loss} for each such sample. 
Thus, RigNet learns to edit each dimension of the control space independently, while also being able to combine the edits using the same network. %
Fig.~\ref{fig:mixing_all} shows mixing results where pose, expression and illumination parameters are transferred from the source to target images.

\section{Limitations}
While we have demonstrated high quality semantic control of StyleGAN-generated facial imagery, our approach is still subject to a few limitations that can be addressed in follow-up work.
In the analysis sections, we have already discussed that StyleRig is not able to exploit the full expressivity of the parametric face model.
This provides a nice insight into the inner workings of StlyeGAN and allows us to introspect the biases it learned.
In the future, this might lead the ways to designing better generative models.
Our approach is also limited by the quality of the employed differentiable face reconstruction network.
Currently, this model does not allow us to reconstruct fine-scale detail, thus we can not explicitly control them.
Finally, there is no explicit constraint that tries to preserve parts of the scene that are not explained by the parameteric face model, e.g., the background or hair style.
Therefore, these parts can not be controlled and might change when editing the parameters.

\section{Conclusion}
We have proposed StyleRig, a novel approach that provides face rig-like control over a pretrained and fixed StyleGAN network.
Our network is trained in a self-supervised manner and does not require any additional images or manual annotations.
At test time, our method generates images of faces with the photorealism of StyleGAN, while providing explicit control over a set of semantic control parameters.
We believe that the combination of computer graphics control with deep generative models enables many exciting editing applications, provides insights into the inner workings of the generative model, and will inspire follow-up work.

\let\thefootnote\relax\footnotetext{
	\textbf{Acknowledgements:}
We thank True-VisionSolutions Pty Ltd for providing the 2D face tracker.
This work was supported by the ERC Consolidator Grant 4DReply (770784), the Max Planck Center for Visual Computing and Communications (MPC-VCC), and by Technicolor.
}

{\small
\bibliographystyle{ieee_fullname}
\bibliography{egbib}

\begin{thebibliography}{10}\itemsep=-1pt

\bibitem{Abdal_2019_ICCV}
Rameen Abdal, Yipeng Qin, and Peter Wonka.
\newblock {Image2StyleGAN}: How to embed images into the stylegan latent space?
\newblock In {\em International Conference on Computer Vision (ICCV)}, 2019.

\bibitem{Alexander:2009}
Oleg Alexander, Mike Rogers, William Lambeth, Matt Chiang, and Paul Debevec.
\newblock The digital emily project: Photoreal facial modeling and animation.
\newblock In {\em ACM SIGGRAPH Courses}, pages 12:1--12:15, 2009.

\bibitem{Blanz2003}
Volker Blanz, Curzio Basso, Tomaso Poggio, and Thomas Vetter.
\newblock Reanimating faces in images and video.
\newblock In {\em Computer graphics forum}, pages 641--650. Wiley Online
  Library, 2003.

\bibitem{Blanz1999}
Volker Blanz and Thomas Vetter.
\newblock A morphable model for the synthesis of {3D} faces.
\newblock In {\em SIGGRAPH's Computer Graphics and Interactive Techniques},
  pages 187--194, 1999.

\bibitem{BoothRPDZ18}
James Booth, Anastasios Roussos, Allan Ponniah, David Dunaway, and Stefanos
  Zafeiriou.
\newblock Large scale {3D} morphable models.
\newblock {\em International Journal of Computer Vision (IJCV)},
  126(2):233--254, Apr. 2018.

\bibitem{Cao:2014}
Chen Cao, Yanlin Weng, Shun Zhou, Yiying Tong, and Kun Zhou.
\newblock Facewarehouse: A {3D} facial expression database for visual
  computing.
\newblock {\em IEEE Transactions on Visualization and Computer Graphics
  (TVCG)}, 20(3):413--425, Mar. 2014.

\bibitem{Cao16}
Chen Cao, Hongzhi Wu, Yanlin Weng, Tianjia Shao, and Kun Zhou.
\newblock Real-time facial animation with image-based dynamic avatars.
\newblock {\em ACM Transactions on Graphics (Proceedings of SIGGRAPH)},
  35(4):126:1--126:12, 2016.

\bibitem{chan2019dance}
Caroline Chan, Shiry Ginosar, Tinghui Zhou, and Alexei~A Efros.
\newblock Everybody dance now.
\newblock In {\em International Conference on Computer Vision (ICCV)}, 2019.

\bibitem{deng2019}
Yu Deng, Jiaolong Yang, Sicheng Xu, Dong Chen, Yunde Jia, and Xin Tong.
\newblock Accurate {3D} face reconstruction with weakly-supervised learning:
  From single image to image set.
\newblock In {\em CVPR Workshops}, 2019.

\bibitem{egger20193d}
Bernhard Egger, William A.~P. Smith, Ayush Tewari, Stefanie Wuhrer, Michael
  Zollhoefer, Thabo Beeler, Florian Bernard, Timo Bolkart, Adam Kortylewski,
  Sami Romdhani, Christian Theobalt, Volker Blanz, and Thomas Vetter.
\newblock 3d morphable face models -- past, present and future.
\newblock {\em arXiv preprint arXiv:1909.01815}, 2019.

\bibitem{GarriZCVVPT2016}
Pablo Garrido, Michael Zollh\"{o}fer, Dan Casas, Levi Valgaerts, Kiran
  Varanasi, Patrick P\'{e}rez, and Christian Theobalt.
\newblock Reconstruction of personalized {3D} face rigs from monocular video.
\newblock {\em ACM Trans. on Graph. (Proceedings of SIGGRAPH)}, 35(3):28:1--15,
  June 2016.

\bibitem{Gatys16}
L.~A. {Gatys}, A.~S. {Ecker}, and M. {Bethge}.
\newblock Image style transfer using convolutional neural networks.
\newblock In {\em CVPR}, pages 2414--2423, 2016.

\bibitem{gecer2018semi}
Baris Gecer, Binod Bhattarai, Josef Kittler, and Tae-Kyun Kim.
\newblock Semi-supervised adversarial learning to generate photorealistic face
  images of new identities from 3d morphable model.
\newblock In {\em Proceedings of the European Conference on Computer Vision
  (ECCV)}, pages 217--234, 2018.

\bibitem{genova2018unsupervised}
Kyle Genova, Forrester Cole, Aaron Maschinot, Aaron Sarna, Daniel Vlasic, and
  William~T Freeman.
\newblock Unsupervised training for 3d morphable model regression.
\newblock In {\em Proceedings of the IEEE Conference on Computer Vision and
  Pattern Recognition}, pages 8377--8386, 2018.

\bibitem{NIPSGood}
Ian Goodfellow, Jean Pouget-Abadie, Mehdi Mirza, Bing Xu, David Warde-Farley,
  Sherjil Ozair, Aaron Courville, and Yoshua Bengio.
\newblock Generative adversarial nets.
\newblock In {\em Advances in Neural Information Processing Systems (NIPS)},
  pages 2672--2680. 2014.

\bibitem{Isola17}
Phillip Isola, Jun-Yan Zhu, Tinghui Zhou, and Alexei~A Efros.
\newblock Image-to-image translation with conditional adversarial networks.
\newblock {\em CVPR}, 2017.

\bibitem{gansteerability}
Ali Jahanian, Lucy Chai, and Phillip Isola.
\newblock On the "steerability" of generative adversarial networks.
\newblock {\em arXiv preprint arXiv:1907.07171}, 2019.

\bibitem{KarraALL2018}
Tero Karras, Timo Aila, Samuli Laine, and Jaakko Lehtinen.
\newblock Progressive growing of {GANs} for improved quality, stability, and
  variation.
\newblock In {\em International Conference on Learning Representation (ICLR)},
  2018.

\bibitem{Karra19}
Tero Karras, Samuli Laine, and Timo Aila.
\newblock A style-based generator architecture for generative adversarial
  networks.
\newblock {\em CVPR}, 2019.

\bibitem{kim2018DeepVideo}
H. Kim, P. Garrido, A. Tewari, W. Xu, J. Thies, N. Nie{\ss}ner, P. P{\'e}rez,
  C. Richardt, M. Zollh{\"o}fer, and C. Theobalt.
\newblock {Deep Video Portraits}.
\newblock {\em ACM Trans. on Graph. (Proceedings of SIGGRAPH)}, 2018.

\bibitem{KimZTTRT17}
Hyeongwoo Kim, Michael Zollh{\"{o}}fer, Ayush Tewari, Justus Thies, Christian
  Richardt, and Christian Theobalt.
\newblock {InverseFaceNet: Deep Single-Shot Inverse Face Rendering From {A}
  Single Image}.
\newblock In {\em CVPR}, 2018.

\bibitem{Li17}
Tianye Li, Timo Bolkart, Michael~J. Black, Hao Li, and Javier Romero.
\newblock {FLAME}: Learning a model of facial shape and expression from 4d
  scans.
\newblock {\em ACM Trans. on Graph.}, 36(6):194:1--194:17, 2017.

\bibitem{Liu17UNIT}
Ming-Yu Liu, Thomas Breuel, and Jan Kautz.
\newblock Unsupervised image-to-image translation networks.
\newblock In {\em NIPS}, pages 700--708. 2017.

\bibitem{Brualla18}
Ricardo Martin-Brualla, Rohit Pandey, Shuoran Yang, Pavel Pidlypenskyi,
  Jonathan Taylor, Julien Valentin, Sameh Khamis, Philip Davidson, Anastasia
  Tkach, Peter Lincoln, Adarsh Kowdle, Christoph Rhemann, Dan~B Goldman, Cem
  Keskin, Steve Seitz, Shahram Izadi, and Sean Fanello.
\newblock Lookingood: Enhancing performance capture with real-time neural
  re-rendering.
\newblock {\em ACM Trans. on Graph. (Proceedings of SIGGRAPH-Asia)},
  37(6):255:1--255:14, 2018.

\bibitem{Park19SPADE}
Taesung Park, Ming-Yu Liu, Ting-Chun Wang, and Jun-Yan Zhu.
\newblock Semantic image synthesis with spatially-adaptive normalization.
\newblock In {\em CVPR}, 2019.

\bibitem{Richardson_2017_CVPR}
Elad Richardson, Matan Sela, Roy Or-El, and Ron Kimmel.
\newblock Learning detailed face reconstruction from a single image.
\newblock In {\em CVPR}, 2017.

\bibitem{RingNet19}
Soubhik Sanyal, Timo Bolkart, Haiwen Feng, and Michael Black.
\newblock Learning to regress {3D} face shape and expression from an image
  without {3D} supervision.
\newblock In {\em CVPR}, pages 7763--7772, 2019.

\bibitem{SaragLC2011}
Jason~M. Saragih, Simon Lucey, and Jeffrey~F. Cohn.
\newblock Deformable model fitting by regularized landmark mean-shift.
\newblock {\em IJCV}, 91(2):200--215, 2011.

\bibitem{sela2017unrestricted}
Matan Sela, Elad Richardson, and Ron Kimmel.
\newblock {Unrestricted Facial Geometry Reconstruction Using Image-to-Image
  Translation}.
\newblock In {\em ICCV}, 2017.

\bibitem{selim16}
Ahmed Selim, Mohamed Elgharib, and Linda Doyle.
\newblock Painting style transfer for head portraits using convolutional neural
  networks.
\newblock {\em ACM Trans. on Graph. (Proceedings of SIGGRAPH)}, pages
  129:1--129:18, 2016.

\bibitem{shen2019interpreting}
Yujun Shen, Jinjin Gu, Xiaoou Tang, and Bolei Zhou.
\newblock Interpreting the latent space of gans for semantic face editing.
\newblock {\em arXiv preprint arXiv:1907.10786}, 2019.

\bibitem{Tewari19FML}
Ayush Tewari, Florian Bernard, Pablo Garrido, Gaurav Bharaj, Mohamed Elgharib,
  Hans-Peter Seidel, Patrick P{\'e}rez, Michael Z{\"o}llhofer, and Christian
  Theobalt.
\newblock Fml: Face model learning from videos.
\newblock In {\em Proceedings of the IEEE Conference on Computer Vision and
  Pattern Recognition}, pages 10812--10822, 2019.

\bibitem{Tewari2020NeuralSTAR}
Ayush Tewari, Ohad Fried, Justus Thies, Vincent Sitzmann, Stephen Lombardi,
  Kalyan Sunkavalli, Ricardo Martin-Brualla, Tomas Simon, Jason Saragih,
  Matthias Nießner, Rohit Pandey, Sean Fanello, Gordon Wetzstein, Jun-Yan Zhu,
  Christian Theobalt, Maneesh Agrawala, Eli Shechtman, Dan~B. Goldman, and
  Michael Zollhöfer.
\newblock {State of the Art on Neural Rendering}.
\newblock {\em Computer Graphics Forum}, 2020.

\bibitem{tewari2018self}
Ayush Tewari, Michael Zollh{\"o}fer, Pablo Garrido, Florian Bernard, Hyeongwoo
  Kim, Patrick P{\'e}rez, and Christian Theobalt.
\newblock Self-supervised multi-level face model learning for monocular
  reconstruction at over 250 hz.
\newblock In {\em CVPR}, 2018.

\bibitem{tewari17MoFA}
Ayush Tewari, Michael Zollh{\"o}fer, Hyeongwoo Kim, Pablo Garrido, Florian
  Bernard, Patrick Perez, and Theobalt Christian.
\newblock {MoFA: Model-based Deep Convolutional Face Autoencoder for
  Unsupervised Monocular Reconstruction}.
\newblock In {\em ICCV}, pages 3735--3744, 2017.

\bibitem{thies2019deferred}
Justus Thies, Michael Zollh{\"o}fer, and Matthias Nie{\ss}ner.
\newblock Deferred neural rendering: Image synthesis using neural textures.
\newblock {\em ACM Transactions on Graphics (TOG)}, 38(4):1--12, 2019.

\bibitem{Tran19}
Luan Tran, Feng Liu, and Xiaoming Liu.
\newblock Towards high-fidelity nonlinear 3d face morphable model.
\newblock In {\em CVPR}, June 2019.

\bibitem{Usman_2019_ICCV}
Ben Usman, Nick Dufour, Kate Saenko, and Chris Bregler.
\newblock Puppetgan: Cross-domain image manipulation by demonstration.
\newblock In {\em The IEEE International Conference on Computer Vision (ICCV)},
  October 2019.

\bibitem{Yu_2019_CVPR}
Ye Yu and William A.~P. Smith.
\newblock {InverseRenderNet}: Learning single image inverse rendering.
\newblock In {\em CVPR}, 2019.

\bibitem{zeiler2012adadelta}
Matthew~D. Zeiler.
\newblock Adadelta: An adaptive learning rate method.
\newblock {\em arXiv preprint arXiv:1212.5701}, 2012.

\bibitem{CycleGAN2017}
Jun-Yan Zhu, Taesung Park, Phillip Isola, and Alexei~A Efros.
\newblock Unpaired image-to-image translation using cycle-consistent
  adversarial networks.
\newblock In {\em ICCV}, 2017.

\bibitem{Zollhoefer2018FaceSTAR}
M. Zollh\"{o}fer, J. Thies, P. Garrido, D. Bradley, T. Beeler, P. P\'{e}rez, M.
  Stamminger, M. Nie{\ss}ner, and C. Theobalt.
\newblock {State of the Art on Monocular 3D Face Reconstruction, Tracking, and
  Applications}.
\newblock {\em Comput. Graph. Forum (Eurographics State of the Art Reports
  2018)}, 37(2), 2018.

\end{thebibliography}
}
\clearpage


\section*{Supplementary material}

# StyleRig: Rigging StyleGAN for 3D Control over Portrait Images
## –Supplementary Material–

Ayush Tewari[1]   Mohamed Elgharib[1]   Gaurav Bharaj[2]   Florian Bernard[1]
Hans-Peter Seidel[1]   Patrick Pérez[3]   Michael Zollhöfer[4]   Christian Theobalt[1]

[1]MPI Informatics, Saarland Informatics Campus   [2]Technicolor   [3]Valeo.ai   [4]Stanford University

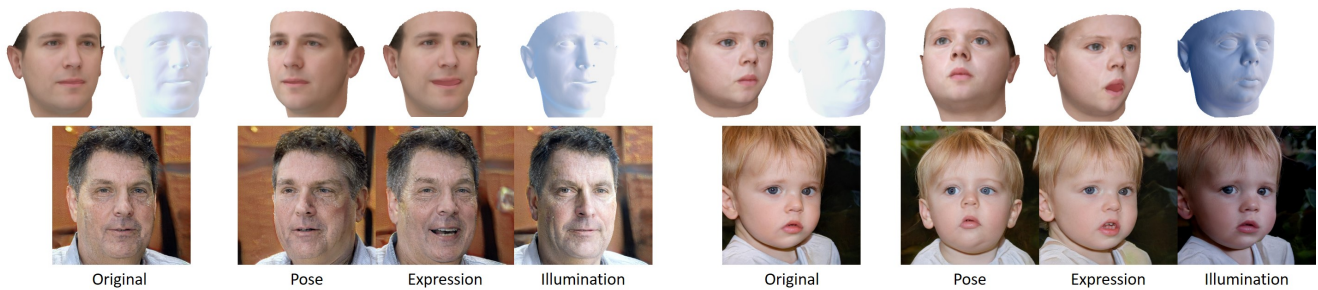

Figure 1: StyleRig allows for face rig-like control over StyleGAN generated portrait images, by translating semantic edits on 3D face meshes to the input space of StyleGAN.

In this supplemental document, we provide further training details and evaluations. We strongly recommend to watch the supplementary video for more editing results.

## 1. Training Details

We use $\lambda_{land} = 17.5$ for pose editing, $\lambda_{land} = 100.0$ for expression editing and $\lambda_{land} = 7.8$ for illumination editing networks. The same hyperparameters are used for both the editing and consistency losses. When we train networks for simultaneous control, we weight the loss functions for the different parameters differently. Rotation losses are weighted by $1.0$, expression by $1000.0$ and illumination by $0.001$. As before, the weights for both the editing and the consistency losses are equal.

We do not edit the translation of the face. We noticed that the training data for StyleGAN was cropped using facial landmarks, such that there is a strong correlation between the head rotation and the translation parameters. Thus, even when training networks to edit other parameters, we do not try to preserve the translation component, for eg., the face is allowed to translate while rotating.

## 2. Evaluation of Simultaneous Parameter Edits

As mentioned in the paper, we can also train networks to edit all three sets of parameters (pose, expression and illumination) simultaneously using a single network. As shown in the results section of the main paper as well as the supplemental video, this produces high quality results. To compare the simultaneous editing performance to networks

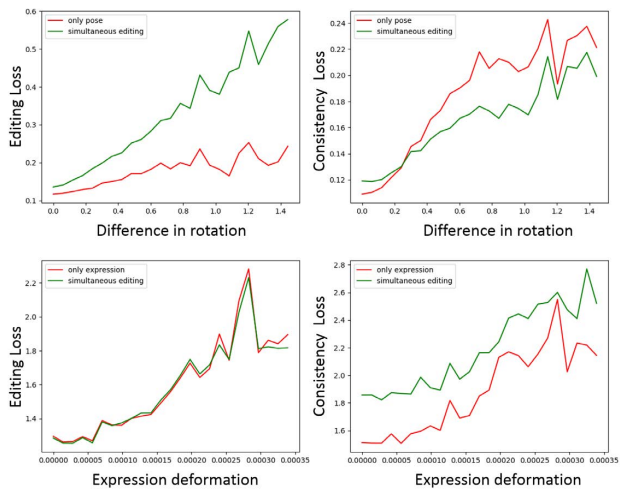

Figure 2: Comparison of models trained to edit individual parameters and the model trained to edit all parameters simultaneously.

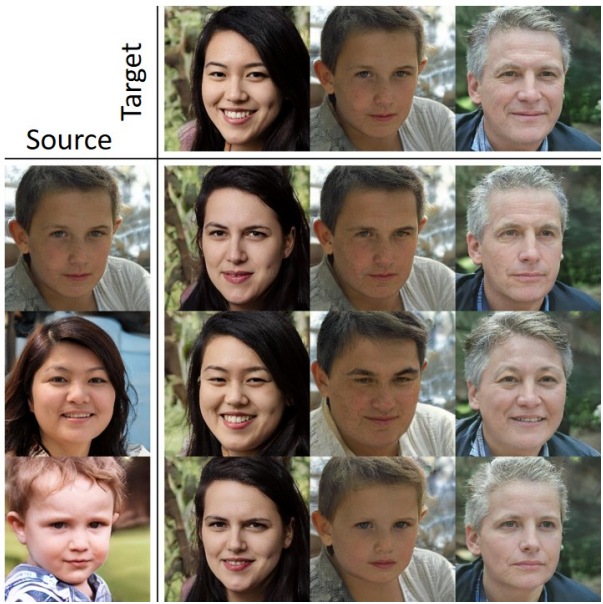

Figure 3: We can also transfer the identity geometry of source images to the target using StyleRig.

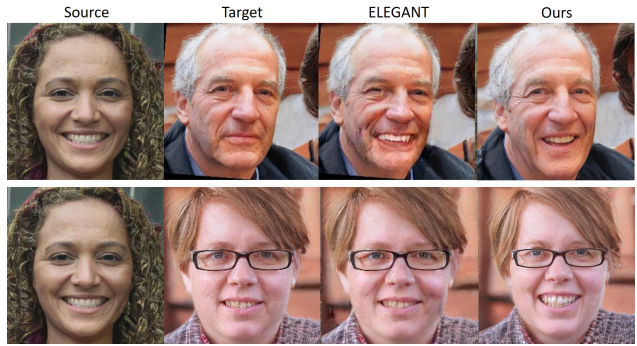

Figure 4: Comparison to ELEGANT [2]. Source expressions are transferred to the target images. We obtain higher quality results, and a better transfer of the source expressions.

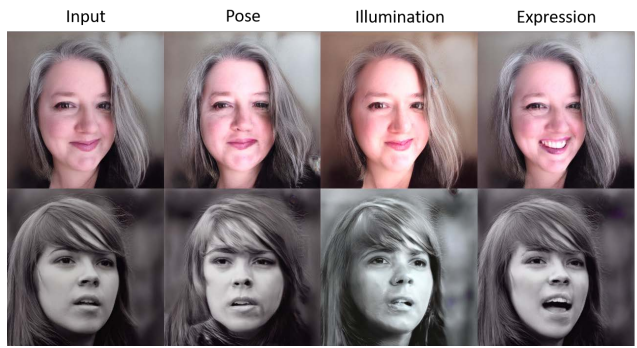

Figure 5: StyleRig can also be used for editing real images. We first optimize the latent embedding of StyleGAN of an input image using Image2StyleGAN [1]. RigNet is then used to edit the result. In some cases such as the bottom row, this leads to artifacts since the optimized latent embedding can be far from the training data.

that have been trained for editing just a single parameter, we plot the editing and consistency losses with respect to the magnitude of edits in Fig. 2. These numbers are computed for 2500 parameter mixing results on a test set. Rotation difference is measured by the magnitude of the rotation angle between the source and target samples in an axis-angle representation. Expression difference is computed as the $\ell_2$ difference between the mesh deformations due to expressions in the source and target samples. All losses are lower when the edits are smaller and increase gradually with larger edits. For the rotation component, the editing loss for the network trained for simultaneous control increases faster. This implies that this network is worse at reproducing the target pose, compared to the network trained only for pose editing. For expressions, while the editing loss remains similar, the consistency losses are higher for the network with simultaneous control. This implies that the network with only expression control is better at preserving other properties (pose, illumination, identity) during editing.

## 3. Geometry Editing

Similar to rotation, expression and illumination, we can also control the identity geometry of faces using the identity component of the 3DMM. Fig. 3 shows several geometry mixing results, where the source geometry can be transferred to the target images.

## 4. Comparison

We compare our approach to ELEGANT [2], a GAN-based image editing approach. Source expressions are transferred to the target images. We obtain higher-quality results with fewer artifacts. We can also better transfer the source expressions to the target.

## 5. Editing Real Images

Our method can also be extended for editing real images. We use the recent Image2StyleGAN approach [1] to compute the latent embedding, given an existing real image. RigNet can then be used to compute the edited embedding, thus allowing for editing high-resolution images, see Fig. 5. However, in some cases, such as Fig. 5 (bottom), this approach can lead to artifacts in the edited results, since the embedding optimized using Image2StyleGAN might be outside the training distribution used for training RigNet.

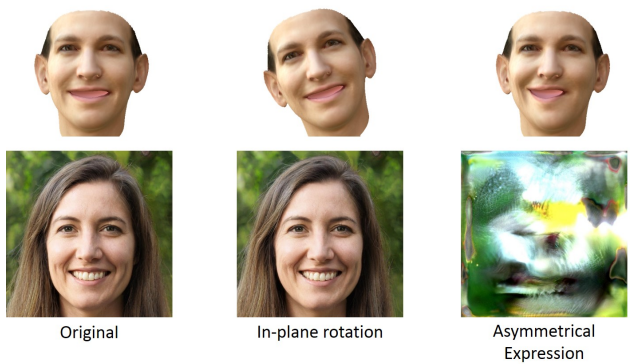

Figure 6: Limitations: Transformations not present in the training data cannot be produced. Thus, our method cannot handle in-plane rotation and asymmetrical expressions.

## 6. Limitations

We show some failure cases in Fig. 6. As explained in the main paper, in-plane rotations can not be produced by our approach. Expressions other than mouth open/smiling are either ignored or incorrectly mapped. As detailed in the main paper, we attribute these problems to a bias in the training data that has been used for training StyleGAN. In addition, we cannot control high-frequency details in the image, since our employed differentiable face reconstruction network only reconstructs coarse geometry and appearance.

## References

[1] Rameen Abdal, Yipeng Qin, and Peter Wonka. Image2StyleGAN: How to embed images into the stylegan latent space? In *International Conference on Computer Vision (ICCV)*, 2019.

[2] Taihong Xiao, Jiapeng Hong, and Jinwen Ma. Elegant: Exchanging latent encodings with gan for transferring multiple face attributes. In *Proceedings of the European conference on computer vision (ECCV)*, pages 168–184, 2018.


\end{document}